\documentclass{article}

\usepackage[nonatbib,preprint]{neurips_2022}

\usepackage[utf8]{inputenc} 
\usepackage[T1]{fontenc}    
\usepackage{hyperref}       
\usepackage{url}            
\usepackage{booktabs}       
\usepackage{amsfonts}       
\usepackage{nicefrac}       
\usepackage{microtype}      
\usepackage{xcolor}         
\usepackage{amsmath}
\usepackage{graphicx}

\title{Learning in Factored Domains with \\ Information-Constrained Visual Representations}
\author{%
  Tyler Malloy\\
  Rensselaer Polytechnic Institute\\
  \texttt{mallot@rpi.edu} \\
  \And
  Tim Klinger \\
  IBM Research AI\\
  \texttt{tklinger@us.ibm.com} \\
  \And
  Miao Liu \\
  IBM Research AI\\
  \texttt{miao.liu1@ibm.com} \\
  \And
  Matthew Riemer \\
  IBM Research AI\\
  \texttt{mdriemer@us.ibm.com} \\
  \And
  Gerald Tesauro \\
  IBM Research AI\\
  \texttt{gtesauro@us.ibm.com} \\
  \And
  Chris R. Sims \\
  Rensselaer Polytechnic Institute\\
  \texttt{simsc3@rpi.edu} \\
}

\begin{document}

\maketitle

\begin{abstract}
Humans learn quickly even in tasks that contain complex visual information. This is due in part to the efficient formation of compressed representations of visual information, allowing for better generalization and robustness. However, compressed representations alone are insufficient for explaining the high speed of human learning. Reinforcement learning (RL) models that seek to replicate this impressive efficiency may do so through the use of factored representations of tasks. These informationally simplistic representations of tasks are similarly motivated as the use of compressed representations of visual information. Recent studies have connected biological visual perception to disentangled and compressed representations. This raises the question of how humans learn to efficiently represent visual information in a manner useful for learning tasks. In this paper we present a model of human factored representation learning based on an altered form of a $\beta$-Variational Auto-encoder used in a visual learning task. Modelling results demonstrate a trade-off in the informational complexity of model latent dimension spaces, between the speed of learning and the accuracy of reconstructions.
\end{abstract}

\section{Introduction}

Deep Reinforcement Learning (DRL) has achieved super-human performance on a variety of tasks by leveraging large neural networks trained on long timescales \cite{mnih2013playing}. However, much of the research in applying RL onto cognitive modelling of human learning has been limited to domains with small state and action sizes \cite{niv2015reinforcement}, due to the low sample efficiency of traditional DRL methods \cite{botvinick2019reinforcement}. 

Recent methods have applied DRL onto predicting human learning by modifying $\beta$-Variational Auto-Encoders ($\beta$-VAE) to additionally predict utility in a supervised fashion \cite{malloy2022modelinghuman}. Disentangled representations have also been applied into improving zero-shot transfer learning in the DRL setting by using latent representations as input to a policy network \cite{higgins2017darla}. The model presented in this work differs from these previous approaches by applying a hypothesis generation and evaluation method onto latent representations, in the context of a factored task representation. 

Factored representations of state transition and reward functions can be used by RL methods to improve generalization and robustness in tasks with a causal structure that corresponds to the factored Markov Decision Process problem specification \cite{kearns1999efficient}. This could be a useful source of higher sample efficiency required to predict human learning using deep learning methods. 

The model presented in this work seeks to leverage the disentangled representations learned by $\beta$-VAE models onto learning the factored representation of a task. This is achieved by generating a set of hypotheses that predict future rewards and states based on the latent features of visual information. This hypothesis space is used to explain the causal structure of a given task, and is repeatedly re-evaluated and re-generated based on the experience of the agent. 

\section{Beta Variational Autoencoders}
The $\beta$-Variational Autoencoder model consists of a deep neural network $q_{\phi}(z|x)$ that learns information-constrained representations of visual information $x$. These representations take the form of a vector of means $\mu_z$ and variances $\sigma_z$ that define a multi-variate Gaussian $\mathcal{N}(\mu_z, \sigma_z)$. This distribution is sampled from to produce a vector of values $z$ that is then fed through the subsequent network layers $p_{\theta}(x|z)$ to produce a reconstruction, the entire model being trained to minimize the difference between the input and reconstruction by maximizing the objective function \cite{burgess2018understanding}: 

\begin{align}
\begin{split}
    \mathcal{L}(\theta,\phi;x,z,\beta) = \mathbb{E}_{q_{\phi}(z|x)}[\log p_{\theta}(x|z)] - \beta  D_{KL}\big( q_{\phi}(z|x) || p(z)\big) 
\end{split}
\end{align}

The $\beta$ parameter allows for additional control over the information bottleneck of the model by adding a weight to the informational complexity of the latent representations defining the multi-variate Gaussian distribution. The result is that the entire model is trained to balance reconstruction accuracy and latent representation complexity in an adjustable fashion. 

\section{Reinforcement Learning for Factored MDPs}
Reinforcement Learning (RL) for Factored MDPs seeks to solve the problem specification described by the Factored Markov Decision Process (FMDP). The FMDP setting is a special case of MDP formed by relating it to a \textit{dynamic Bayesian network} defined by a directed acyclic graph $G_T$ with nodes $\{X_1, X_2, ... , X_n \}$ and scopes $S_1, ..., S_n$ \cite{kearns1999efficient}. A scope $S_i$ of this network describes the dependencies of future state features or rewards based on previous features and actions, with $x[S_i]$ signifying the features of state $x$ corresponding to the scope $S_i$. This allows for a definition of the factored transition function $P(x'|x,a)$ and reward function $R(x)$ as follows \cite{sallans2004reinforcement}: 
\begin{equation}
\begin{split}
P(x'|x,a) = \prod_{i=1}^n P_i(x_i'|x[S_i],a) \quad  \quad R(x) = \dfrac{1}{n} \sum_{i=1}^{n} R_i(x[S_i])
\end{split} \label{Eq:factored}
\end{equation}

These factored representations can be leveraged to significantly improve sample efficiency when the causal structure is provided \cite{chen2020efficient}. However, it can be difficult to learn these factored representations from scratch, especially in environments with complex information such as visual domains. In the following section we describe how the proposed model leverages disentangled latent representations with a given hypothesis generation method to produce useful factored representations. 

\section{Proposed Model}
\begin{figure}[t!] 
\begin{centering}
  \includegraphics[width=12cm]{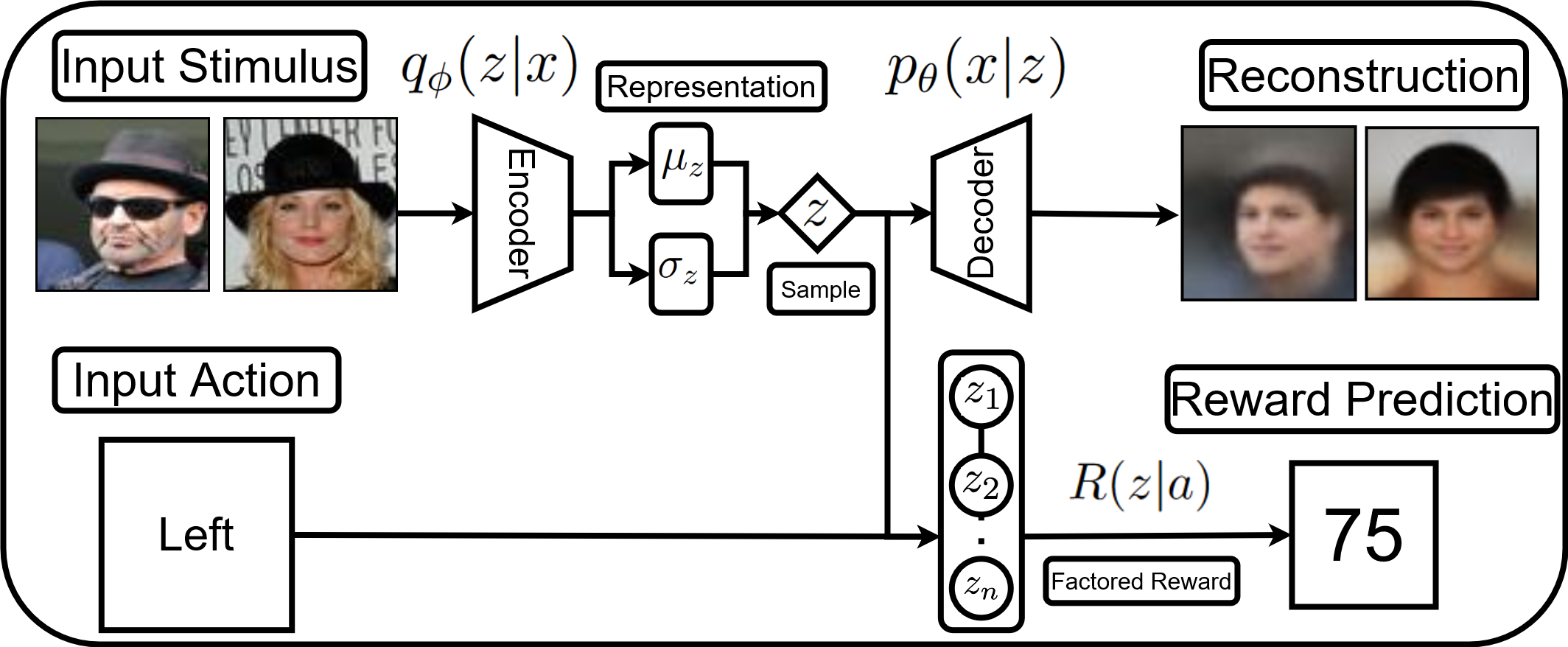} 
  \caption{Example of the RL$\beta$-VAE model forming a reconstruction and predicted reward.}\label{fig:Model}
 \end{centering} 
\end{figure}

The proposed RL$\beta$-VAE model (see Figure \ref{fig:Model}) begins with a slight alteration to the $\beta$-VAE, in order to additionally make predictions of the reward associated with a stimuli and action pair. The resulting network is trained with the following objective:
\begin{align}
\begin{split} 
    \mathcal{L}(\theta,\phi;x,z,\beta, \upsilon, r) = \mathbb{E}_{q_{\phi}(z|x)}[\log p_{\theta}(x|z)] - \beta  D_{KL}\big( q_{\phi}(z|x) || p(z)\big) + \upsilon \big(R(z|a) - r \big)^2  
\end{split} 
\end{align}
where $\upsilon$ is an additional parameter that weighs the importance of the accuracy of reward predictions and the reward $R(z|a)$ is defined in terms of the factored reward of the latent representation $Z$, and the discounted value of the subsequent latent representation $Z'$ observed after performing action $a$:
\begin{equation}
\begin{split}
\quad R(z|a) = \dfrac{1}{n} \sum_{i=1}^{n} R_i(z[S_i]) + \gamma V\bigg(\prod_{i=1}^n P_i(z_i'|z[S_i],a)\bigg)
\end{split}
\end{equation}
Where $\gamma V(Z')$ is the discounted value of the subsequent latent representation $Z'$, here calculated using the factored transition function from Eq. \ref{Eq:factored}. This model uses unsupervised pre-training using a reward of 0 to calculate the training loss. After pre-training, the model can leverage the learned disentangled representations to predict a factored reward structure that allows for improved generalization and robustness, resulting in higher sample efficiency. 

To transition from disentangled latent features to a factored representation requires the generation and evaluation of a set of hypotheses that correspond to potential scopes $S_1, ..., S_n$ . The method of hypothesis generation and evaluation used here has been previously applied onto abstract inductive reasoning \cite{tenenbaum2006theory}. The steps of this process consist of 1) sampling a reduced hypothesis space $\mathcal{H}^* \subseteq \mathcal{H}$ from a probability distribution $q(\mathcal{H}^*)$ and 2) evaluating the hypotheses in the reduced space through some metric for how well the hypothesis matches experience \cite{bonawitz2010deconfounding}. For an example of the factored hypothesis generation and evaluation method see the appendix. 

For the learning task described in this paper, the generation of hypotheses can be achieved through a simple linear fitting of the learned representations to the observed reward. The space of hypotheses consists of all possible scopes $S_1, ..., S_n$ that define the factored reward function. The evaluation step ranks each hypothesis based on mean-squared error of reward prediction accuracy. Alternatives to this approach (including Bayesian inference or TD-error update) are possible, but not required due to the simple structure of the deterministic contextual bandit learning described in the next section. 

\section{Learning Task}
While factored MDPs can aid in the sample efficiency of RL algorithms in many domains, in this learning task we focus on reward factorization using a simple bandit learning environment. This learning task consists of a contextual N-armed bandit based on two images of celebrity faces \cite{liu2015faceattributes}.

The two actions available in the 2-armed bandit setting correspond to selecting the left and right stimuli, meaning we can further simplify the input to the RL$\beta$-VAE model as only the face corresponding to the action chosen. The result is two reward predictions $[r_{\text{left}},r_{\text{right}}]$ which are the input to a simple soft-max function, a method commonly used in cognitive modelling of human bandit learning \cite{niv2015reinforcement}. 

In our contextual bandit task, faces wearing glasses are worth 25 points, wearing hats are worth 50 points, wearing both are worth 75 points and wearing neither are worth 0 points. The assumption of the hypothesis generation method used by the RL$\beta$-VAE  model is that the reward can be predicted by the sum of simple linear functions which map the latent representation values $Z: \{z_0, z_1, ..., z_n\}$ onto the observed reward. As noted previously, more complex hypothesis generation and evaluation methods are possible, but unnecessary for this learning task. 

Before applying the RL$\beta$-VAE  models onto predicting reward they were pre-trained on 100 epochs of the full 220K image dataset of celebrity faces \cite{liu2015faceattributes}, with 100 test images removed. During contextual bandit model testing, two images of celebrities are randomly chosen from a set of 100 (25 each of hats, glasses, both, and neither) images not included in the initial model pre-training. To ensure that one of the options always has a higher reward, the images are selected from different categories. 

\section{Modelling Results}
\begin{figure}[h] 
\begin{centering}
  \includegraphics[width=\columnwidth]{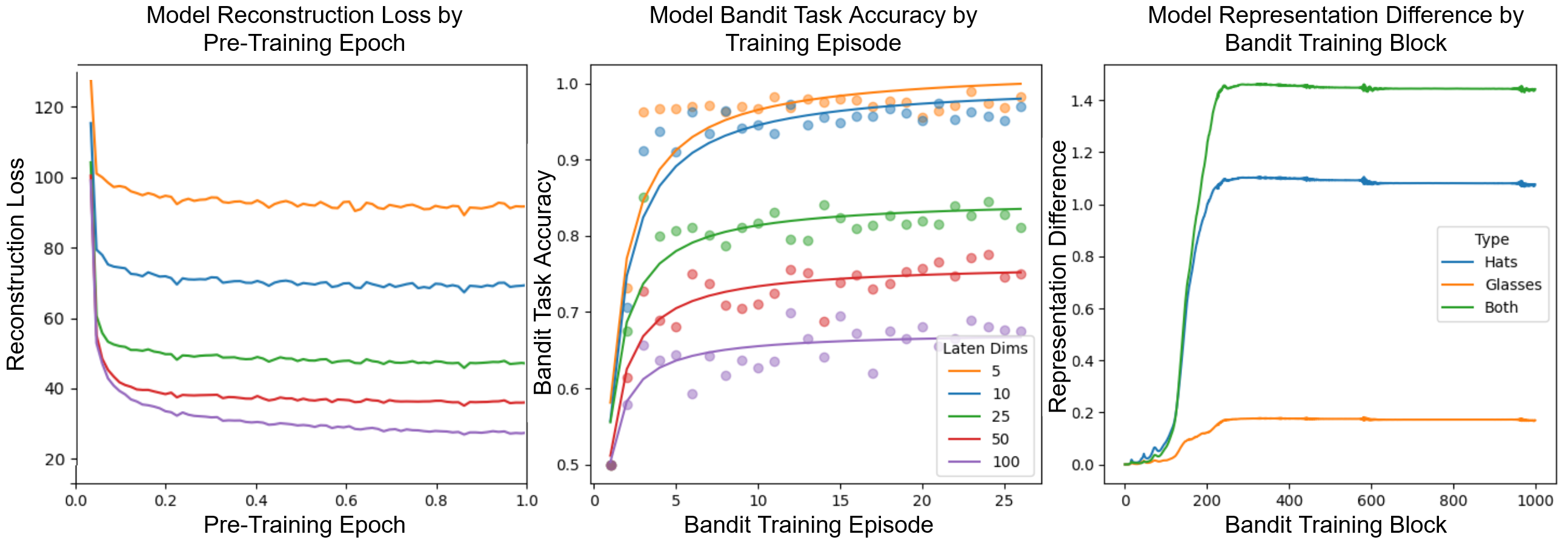}
  \caption{\textbf{Left:} Model pre-training reconstruction loss by training epoch, lower is better, color indicates latent dimension size. \textbf{Middle:} Contextual bandit training for 1000 runs of model accuracy by trail means (dots) are fit to a logarithmic function (lines). \textbf{Right:} Representation difference in mean-squared error between images containing hats, glasses, and both, compared to wearing neither.}\label{fig:Results}
 \end{centering} 
\end{figure}

The main method of assessing the speed of learning in the contextual bandit task is the probability the model assigns to selecting the higher reward bandit arm. The results shown in the middle column of Figure \ref{fig:Results} demonstrate that smaller latent dimension spaces allow for faster learning of the factored reward structure in this contextual bandit task. Notably, the models with small latent dimension sizes are able to consistently select the option with a higher reward after only two experiences in this task.

The left column of Figure \ref{fig:Results} compares reconstruction loss by pre-training epoch. These results demonstrate a lower end of training reconstruction accuracy from models with smaller latent spaces. While these small latent dimensions are useful for quick hypothesis generation, they make accurate reconstruction of stimuli more difficult due to the tight information-bottleneck imposed on the model. 

This represents a trade-off between learning speed and reconstruction accuracy that has direct implications on how the human mind forms constrained representations of visual information that is used in learning tasks. Future research in this area can investigate the specific balance of this trade-off made by humans engaged in learning tasks based on visual information. 

In the right column of Figure \ref{fig:Results}, we compare the average latent representation difference, as measured by mean squared error, between each of the three non-zero utility stimuli types (glasses, hats, both) and the stimuli wearing neither glasses nor hats. Initially all representations are equally similar to stimuli without hats or glasses. As utility is learned, representations of higher utility stimuli become relatively more differentiated. In these results, the low utility stimuli is most similar to the zero utility stimuli, and the highest utility stimuli is most different. This demonstrates a utility-based \textit{acquired equivalence} whereby stimuli with similar utility outcomes have similar latent representations.

\section{Conclusions}
The results presented in this work show the value of disentangled representations of visual information in learning factored rewards. The learning task used in testing these models, while simple, revealed potential explanations of how the human mind performs fast learning through hypothesis generation in an information-compressed space that allows for better generalization and robustness. The method of generating potential hypotheses that explain the reward observed in this contextual bandit task was designed for the deterministic nature of the contextual bandit task, but simple adjustments are possible to extend this application into alternative domains. 

In addition to providing insight into the structure of visual information as it is being processed by the reinforcement learning faculty of the human brain, this work is also related to the question of how best to define disentanglement, which has been identified as an interesting open question \cite{higgins2018towards}. Specifically, the results provided here suggest the usefulness of a behavioural definition of disentanglement, which is achieved when representations are disentangled in a way that makes them useful for behavioural goals such as forming hypotheses that explain experience and direct future behaviour. 


\bibliographystyle{plain} 
\bibliography{infocog} 

\newpage

\section{Appendices}
\subsection{Stimuli examples}
\begin{figure}[h] 
\begin{centering}
  \includegraphics[width=\columnwidth]{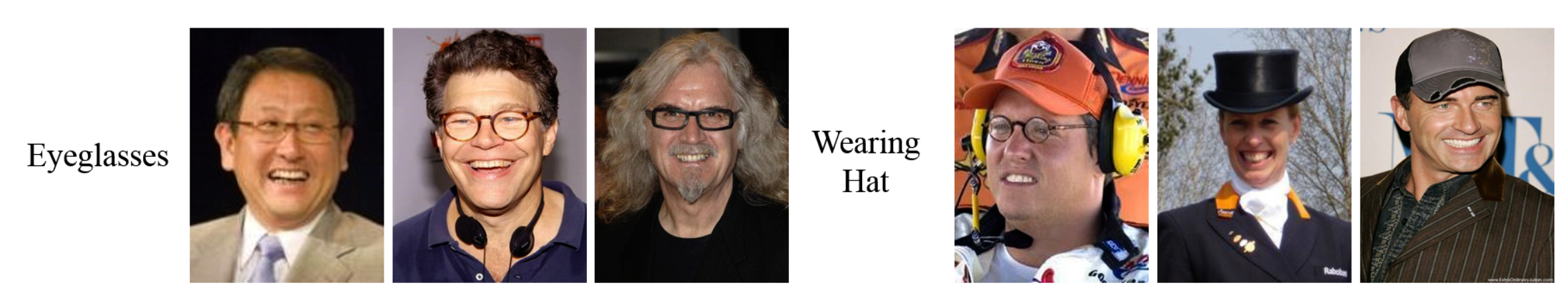}
  \caption{Examples of face images with either eyeglasses or hats from the celebA dataset \cite{liu2015faceattributes}.}\label{fig:GlassesHats}
 \end{centering} 
\end{figure}

\subsection{Hypothesis Generation and Evaluation} \label{Appendix:GenEval}

As mentioned previously the steps of this process consist of 1) sampling a reduced hypothesis space $\mathcal{H}^* \subseteq \mathcal{H}$ from a probability distribution $q(\mathcal{H}^*)$ and 2) evaluating the hypotheses in the reduced space through some metric for how well the hypothesis matches experience \cite{bonawitz2010deconfounding}.

\begin{figure}[h] 
\begin{centering}
  \includegraphics[width=\columnwidth]{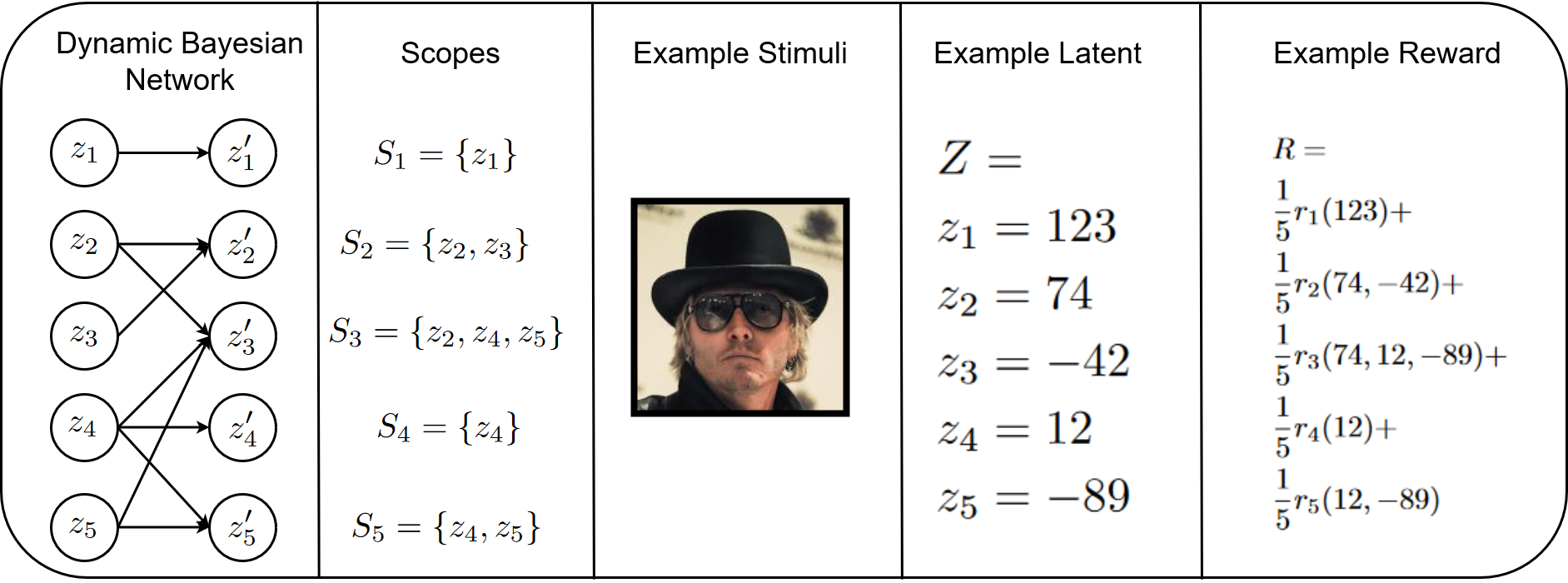} 
  \caption{Example a dynamic Bayesian network defined by one hypothesized scope. An example stimuli with latent representation and factored reward function. Note that the hypothetical DBN describes the transition function which is not used for the contextual bandit task.}\label{fig:Example}
 \end{centering} 
\end{figure}

In the factored MDP setting, a hypothesis is a set of scopes $S_1, ..., S_n$ that correspond to the causal structure of an environment. Figure \ref{fig:Example} shows one possible hypothesis for the causal structure of a learning environment. In this example the first scope $S_1 = \{z_1\}$ corresponds to the relationship between the features contained in $z_1$ for the factored state transition function and reward function described in Eq. \ref{Eq:factored}. This relationship is signified in the Dynamic Bayesian Network in the left column of Figure \ref{fig:Example} by the arrow from $z_1$ to $z'_1$. Because the first scope $S_1$ only contains the feature $z_1$, the first function of the factored reward $r_1$ depends only on the first latent feature $z_1 = 123$. 

The full hypothesis space for the reward of a given latent representation $Z$ of size $n$ with k scope elements is $Z_n^k$ for each of the possible scopes $S_1, ..., S_n$. In the example hypothesis shown in Figure \ref{fig:Example}, n = 5 and k is 1, 2, or 3 and the hypothetical scope is defined as $S_1 = \{z_1\}$, $S_2 = \{z_2, z_3\}$, $S_3 = \{z_2, z_4, z_5\}$, $S_4 = \{z_4\}$, $S_5 = \{z_4, z_5\}$

In practice when performing the contextual bandit task described in the paper, the reduced hypothesis space is formed by selecting some limited complexity of scopes, set as $k = 1$ or $2$, meaning only 1 or 2 elements were contained in each scope, which significantly reduces the possible hypothesis space. 

The probability function sampling the reduced space $q(\mathcal{H}^*)$ was defined to deterministically select the most likely hypothesis as evaluated by the mean-squared error of the most recent reward prediction. This simple evaluation and hypothesis sampling approach was adequate for the deterministic reward setting of this contextual bandit, but a more complex sampling approach is also possible. 

\end{document}